\documentclass[letterpaper]{article}
\usepackage[preprint]{aaai2027}

\usepackage[hyphens]{url}
\usepackage{graphicx}
\usepackage{natbib}
\usepackage{caption}
\usepackage{booktabs}
\usepackage{amsmath}
\usepackage{amssymb}

\newcommand{\D}{\mathcal{D}}
\newcommand{\I}{\mathcal{I}}

\newcommand{\E}{\mathbb{E}}

\newcommand{\sg}{\operatorname{sg}}
\newcommand{\pool}{\operatorname{pool}}
\newcommand{\render}{\operatorname{render}}
\newcommand{\concat}{\operatorname{concat}}
\newcommand{\vts}{\mathrm{vts}}
\newcommand{\orig}{\mathrm{orig}}
\newcommand{\sft}{\mathrm{sft}}

\newcommand{\sgloss}{\mathrm{sg}}
\newcommand{\prmlp}{\mathrm{prmlp}}

\title{When Prompts Become Pixels: Prompt-Region\\Grounding for Multimodal Reasoning}
\author{
Yongxin Wang\textsuperscript{\rm 1},
Ruizhe Zhou\textsuperscript{\rm 2},
Yueling Tang\textsuperscript{\rm 2},
Yingying Zhu\textsuperscript{\rm 3},\\
Xuemin Zhao\textsuperscript{\rm 3},
Xiaojun Chang\textsuperscript{\rm 1,\rm 4},
Xiaodan Liang\textsuperscript{\rm 1,\rm 2}
}
\affiliations{
\textsuperscript{\rm 1}Mohamed bin Zayed University of Artificial Intelligence\\
\textsuperscript{\rm 2}Sun Yat-sen University \quad
\textsuperscript{\rm 3}Transsion \quad
\textsuperscript{\rm 4}University of Science and Technology of China
}

\begin{document}

\maketitle

\begin{abstract}
Multimodal large language models increasingly reason over screenshots and
documents where the task itself may be written in pixels. Yet benchmarks
usually place questions in text, leaving it unclear whether models use the same
instruction equally well across channels. We introduce Visualized Task
Semantics (VTS), a controlled intervention that moves the question into the
image while keeping the source problem and answer fixed. Across six MLLMs and
four benchmarks, accuracy drops in all 24 model-task pairs, by 17.8 points on
average. Models often transcribe the visual question correctly yet fail to use
it, exposing a semantic channel gap beyond OCR. To reduce this gap, we present
prompt-region grounding, whose core design aligns the question region with
typed semantics and recovers its clean representation from a masked view. At
matched training cost, our method raises four-benchmark VTS accuracy from 58.0
to 66.3 while preserving accuracy on the original interface, and requires no
OCR or region metadata at inference. Reading task-bearing text and grounding it
as an instruction for reasoning are distinct capabilities.
\end{abstract}

\section{Introduction}

Visual text has become a standard input to multimodal large language models
(MLLMs). High-resolution models can read screenshots, documents, and other
text-rich images with increasing accuracy
\citep{wang2024qwen2vl,bai2025qwen25vl,wu2024deepseekvl2}. Most work treats
this text as evidence to extract: a label, value, or paragraph contributes
content to the answer. A visual question has a different function. It
determines how the evidence should be used and what answer is required. Whether
MLLMs preserve this task-defining function when a question moves from prompt
tokens into image pixels remains less understood.

Standard reasoning protocols make this function difficult to study. They place
the question in the language channel and reserve the image for visual evidence.
Prior work reports weaker performance when instructions are visualized
\citep{li2024textimages,an2025voqa,liu2026vistabench}, but the measured gap can
mix several effects: canvas expansion, image resizing, renderer errors, missing
prompt content, and the change of semantic channel itself. A model may also
transcribe the visual question correctly without using it to control the
answer. An accuracy drop alone therefore cannot tell us whether the model
failed to read the question or failed to use it.

We study this problem with \emph{Visualized Task Semantics} (VTS), a paired
intervention that changes where the question is presented. Given a benchmark
item with question $q$, image $x$, and answer $a$, the Original view presents
$(q,x)$. VTS renders $q$ above $x$ and replaces the typed question with a fixed
cue. The source problem, visual evidence, and target answer remain paired
within each item. VTS then adds separate controls for canvas geometry, prompt
duplication, and missing task content.

\begin{figure*}[t]
\centering
\includegraphics[width=\textwidth]{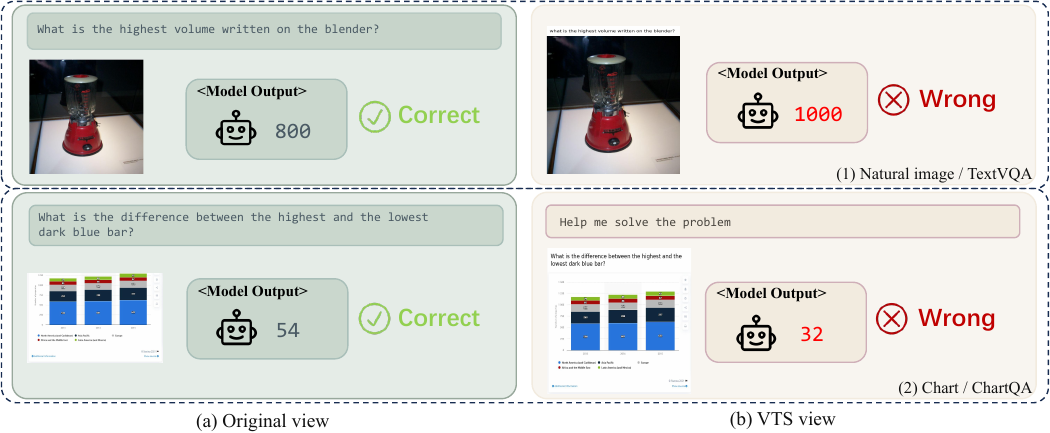}
\caption{The task stays fixed, but the answer changes when its question
becomes pixels. VTS renders the question inside the image and replaces the
native prompt with a fixed cue. The upper schematic omits this separate cue
for space; the lower example displays it. The source visual evidence and
target answer are unchanged within each pair.}
\label{fig:vts_problem_setup}
\end{figure*}

VTS reveals a systematic semantic channel gap. Across six MLLMs and four
benchmarks, accuracy falls in all 24 model-task pairs, by 17.8 points on
average and by as much as 28.0 points. A blank-canvas control remains within
1.1 points of Original accuracy, and a duplicate-question control remains
within 0.6 points. Removing the question from both channels causes accuracy to
collapse. The loss therefore cannot be explained by the added canvas, image
resizing, or renderer corruption alone.

\begin{figure}[t]
\centering
\includegraphics[width=\columnwidth]{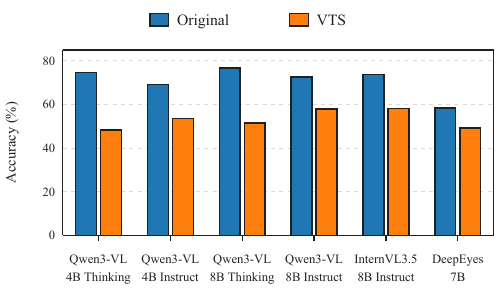}
\caption{Every evaluated model is weaker when the question is in the
image. Bars show mean accuracy over MATH-Vision, MathVista, ChartQA, and MMMU.
Both interfaces are shown because a small gap can also occur when performance
is low.}
\label{fig:performance_findings}
\end{figure}

We next test whether the remaining gap reduces to text recognition. In a
two-task diagnostic, the base model transcribes 87.6\% of visual questions
exactly but answers only 48.4\% correctly. Reinserting the same model's
transcript into the text channel raises answer accuracy by 7.7 points while
leaving the composite unchanged. A substantial part of the gap therefore
appears after transcription: the model recovers the words, but those words
exert less control as pixels than as prompt tokens.

This finding motivates \emph{prompt-region grounding}. Mixed-interface replay
presents questions in both channels, but it does not explicitly connect the
visual question region to its typed counterpart. Our method adds two
region-level objectives. PVRD-SG aligns the prompt region with a frozen
representation of the typed question. PRMLP masks part of that region and
recovers the representation of its clean crop. Component controls test both
parts of this design: prompt-panel readouts outperform full-image and
random-region summaries, while matched targets outperform mismatched targets.
Region boxes and crops are used only during training. At inference, the model
receives one composite and answers directly, without OCR or localization
metadata.

\begin{figure*}[t]
\centering
\includegraphics[width=\textwidth]{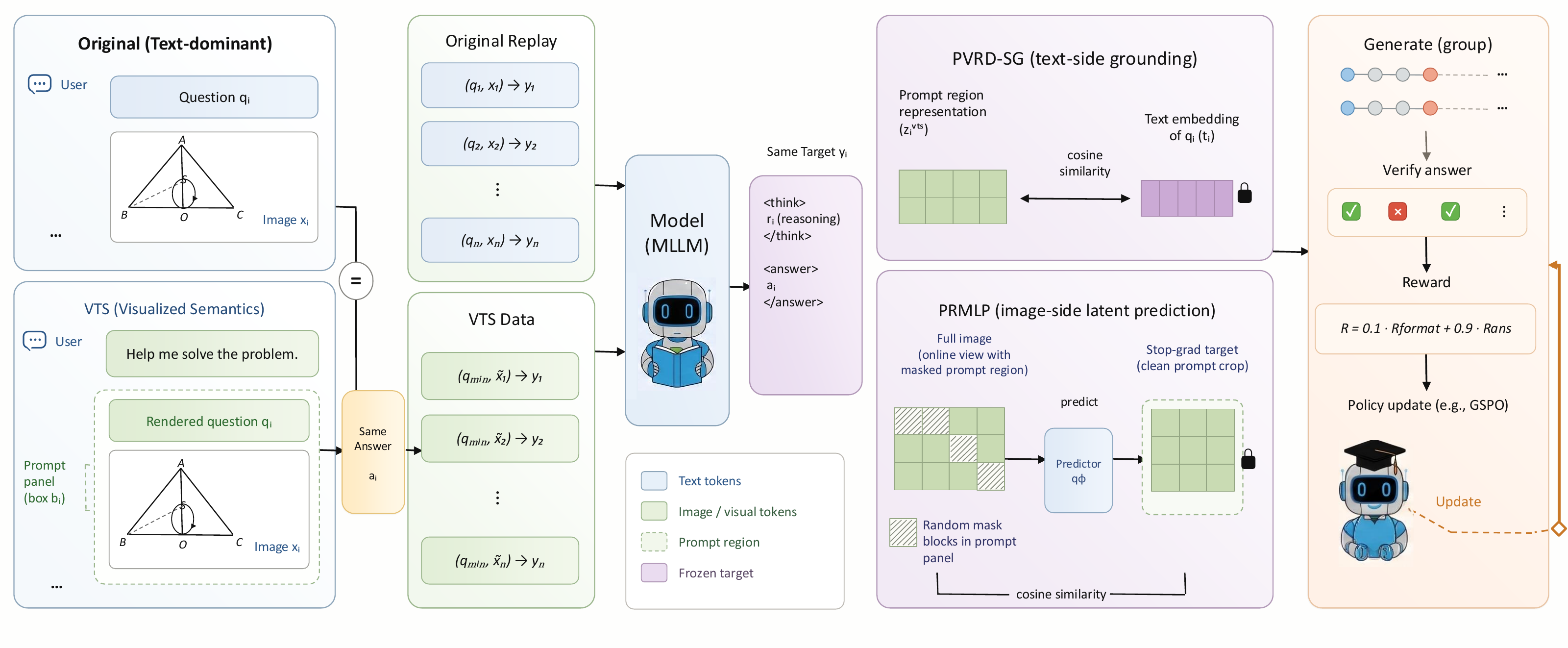}
\caption{Prompt-region grounding. Paired replay exposes the model to
Original and VTS views with the same answer target. PVRD-SG connects the
visual question region to its typed semantics, and PRMLP reconstructs a clean
latent target from a masked prompt region. GSPO then continues the grounded
checkpoint using verifiable format and answer rewards.}
\label{fig:method_overview}
\end{figure*}

At matched training cost on Qwen3-VL-4B, ordinary SFT reaches 58.0 VTS
accuracy and 69.1 Original accuracy across four benchmarks. Prompt-region
grounding raises these scores to 66.3 and 70.3, with a VTS gain on every task.
The same recipe improves a second backbone. It also gains 4.1 points on the
independently constructed VISTA-Bench protocol and 7.9 points on 1,000 held-out
real-world pages.

VTS and prompt-region grounding distinguish two capabilities that MLLM
evaluations often treat as the same: reading text in an image and grounding
that text as the task instruction. VTS measures the distinction under paired
problems, and the region-level objectives train for task use without changing
the inference interface. The results show that the channel carrying a question
can alter reasoning even when its wording and answer remain fixed.

\begin{figure}[t]
\centering
\includegraphics[width=\columnwidth]{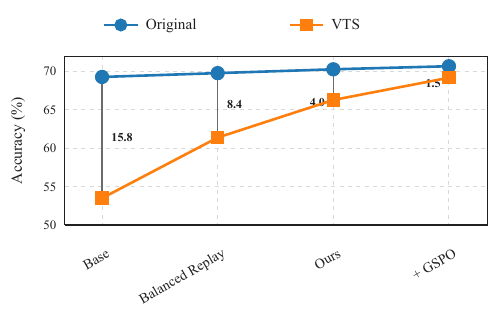}
\caption{Training progressively closes the semantic channel gap.
Lines show four-benchmark mean accuracy. Vertical connectors mark the
Original minus VTS gap at each stage: 15.8, 8.4, 4.0, and 1.5 points.}
\label{fig:adaptation_progression}
\end{figure}

\section{Related work}
\label{sec:related}

\textbf{Multimodal models and reasoning.}
High-resolution MLLMs improve perception in images and documents \citep{wang2024qwen2vl,bai2025qwen25vl,wu2024deepseekvl2,li2024llavaonevision}. Other work improves multimodal reasoning through post-training or by varying diagrams, problem instances, and image count \citep{zhu2025internvl3,kimi2025kimivl,zhang2024mathverse,zou2024dynamath,wang2025mvmath}. Our study holds the source problem fixed and asks how training should handle a change in the interface carrying its question.

Recent work trains multimodal reasoning with process rewards, contrastive
reflection, and verifiable rewards
\citep{wang2025visualprm,wang2025care,guo2025deepseekr1}. We instead change
where the task appears while keeping the answer fixed. We apply GSPO after the
supervised model. The combination of PVRD-SG and PRMLP is the supervised method
contribution; GSPO is the final reinforcement-learning stage.

\textbf{Visual prompting and questions.}
VIM introduces visual-modality instructions, measures the text-to-pixel gap across eight benchmarks, and trains on text and pixel interfaces \citep{li2024textimages}. VoQA places the scene and question in one image and reconstructs the visual question before answering during supervised fine-tuning \citep{an2025voqa}. VISTA-Bench compares matched text and visualized-text questions across renderers and OCR systems \citep{liu2026vistabench}. These works establish the problem; we do not claim that moving a question into pixels is new.

VIM-style mixed replay is therefore a central baseline rather than an omitted
alternative. VoQA is the closest reconstruction-based method: it generates an
intermediate question target before answering. We instead test whether
region-level training can improve direct answering without an output-side
transcription stage. VTS supplies source-matched reasoning pairs for this
comparison. OCRBench v2 and MMLongBench-Doc cover broader visual-text
localization and document layouts
\citep{fu2025ocrbenchv2,ma2024mmlongbenchdoc}. VTCBench studies a related
modality gap for long text compressed into dense images
\citep{zhao2025vtcbench}.

\textbf{Alignment and consistency learning.}
RegionCLIP learns open-vocabulary links between image regions and text \citep{zhong2021regionclip}. Align-KD matches modalities to compress MLLMs, while Align-TI uses a teacher to transfer instruction-relevant visual interactions \citep{feng2025alignkd,chen2026alignti}. These methods do not target printed questions, but they provide close precedents for cross-modal distillation. Masked representation learning provides a second precedent. I-JEPA predicts
masked image regions. BYOL updates a stop-gradient target with an exponential
moving average, whereas SimSiam stops the target gradient without that average
\citep{assran2023selfsupervised,grill2020bootstrap,chen2021exploring}. PRMLP
instead predicts a detached clean-crop representation from a composite whose
prompt region has been masked. The reported implementation uses an identity
predictor.


\section{Prompt-region grounding}
\label{sec:method}

Paired replay teaches both interfaces but leaves their relationship implicit.
Prompt-region grounding makes that relationship an explicit training signal.
It adds two losses on VTS and real-world visual-prompt examples:
\emph{prompt-visual representation distillation with semantic grounding}
(PVRD-SG) connects the visual question region to a frozen representation of
its typed semantics, and \emph{prompt-region masked latent prediction} (PRMLP)
connects a masked question region to its clean visual representation. Both
losses share the original answer target and leave the inference interface
unchanged.

\subsection{Visualized task semantics}

Let $(q_i, x_i, a_i)$ denote a question, its visual context, and its canonical
answer. The original benchmark view is
\begin{equation}
    u_i^{\orig} = \concat(q_i, x_i).
\end{equation}
The model receives $q_i$ as language tokens and $x_i$ through the vision
encoder. VTS moves $q_i$ into the image. A deterministic renderer $\render$
normalizes lightweight LaTeX, wraps the question to the image width, and draws
it in a white panel above the source image:
\begin{equation}
    \tilde{x}_i = \render(q_i, x_i), \qquad
    u_i^{\vts} = \concat(q_{\min}, \tilde{x}_i).
\end{equation}
Here $q_{\min}$ is the fixed cue ``Help me solve the problem''. The renderer records the panel box $b_i$ and saves a clean crop $c_i$. Both views use the same answer $a_i$.

VTS changes both the question channel and the remaining text prompt. We use
three controls to interpret this change. Canvas presents
$\concat(q_i,\render(\varnothing,x_i))$: the native question plus the same
padded canvas with a blank panel. Duplicate presents
$\concat(q_i,\render(q_i,x_i))$, placing the question in both channels.
Image-only presents $\concat(q_{\min},\render(\varnothing,x_i))$, so neither
channel contains the task wording. Renderer template, source-image placement,
decoding, and scorer otherwise remain fixed. The controlled training subset
uses only the Original and VTS views.

During supervised training, the VTS renderer supplies $b_i$ and $c_i$;
the GLM-OCR pipeline supplies the corresponding region and crop for real-world rows
\citep{duan2026glmocr}. At VTS inference, the model receives only the image and
$q_{\min}$. No evaluation supplies a box, crop, OCR transcript, or other
localization metadata. The method uses known regions as supervision, but it is
not an explicit task-region detector.

\subsection{Balanced replay}

Following VIM's cross-interface mixture \citep{li2024textimages}, the
controlled subset requires one Original and one VTS row for every selected
source item:
\begin{equation}
    \begin{aligned}
    \D_{\mathrm{pair}}
    &= \left\{(u_i^{\orig},y_i)\right\}_{i\in\I}
       \cup \left\{(u_i^{\vts},y_i)\right\}_{i\in\I}.
    \end{aligned}
\end{equation}
The complete supervised pool is
$\D_{\sft}=\D_{\mathrm{pair}}\cup\D_{\mathrm{rw}}$, where
$\D_{\mathrm{rw}}$ pairs held-in real-world images with their original prompts
and extracted regions. Let $\I_{\mathrm{vis}}$ index the VTS and real-world
visual-prompt rows; each stores $(q_i,b_i,c_i)$.
We verify the one-to-one construction within $\D_{\mathrm{pair}}$ using stable
pair IDs and target and answer hashes. Both views use the same assistant
target:
\begin{equation}
    y_i =
    \begin{cases}
    s_{\mathrm{ans}}(a_i), & \text{answer-only},\\
    s_{\mathrm{trace}}(r_i)\mathbin{\Vert}s_{\mathrm{ans}}(a_i),
        & \text{trace-bearing}.
    \end{cases}
\end{equation}
Here $r_i$ is a question-scrubbed reasoning trace and $a_i$ is the canonical
answer. The fixed serializers add the corresponding \texttt{<think>} and
\texttt{<answer>} tags. The target policy is frozen before training and shared
by both views. Trace-bearing targets omit verbatim reconstructions of $q_i$;
otherwise the run uses the answer-only target.
The supervised loss is autoregressive negative log-likelihood over assistant
tokens:
\begin{equation}
    \mathcal{L}_{\sft}(\theta) =
    \E_{(u,y) \sim \D_{\sft}}
    \ell_{\mathrm{NLL}}(\theta;u,y),
\end{equation}
where $\ell_{\mathrm{NLL}}$ is the standard autoregressive loss on assistant-target positions.

\subsection{PVRD-SG: Prompt-region semantic grounding}

Balanced replay exposes the model to both interfaces but does not explicitly
tie the rendered question to its typed counterpart. PVRD-SG places this
constraint on the prompt region rather than the whole image. The distinction
is important because a VTS composite contains two kinds of information: the
question specifies the task, while the source image supplies the evidence. A
whole-image summary mixes these roles and may be dominated by objects, charts,
or diagrams outside the question panel. Reading only from the recorded prompt
box asks a narrower question: does the region that contains the instruction
represent the same task as its typed counterpart? Let
$P_i^{\mathrm{text}}$ be a fixed text-only template for $q_i$, and let $Q_i$
contain its question-token positions. Before training, we compute and freeze
\begin{equation}
    t_i=\operatorname{normalize}\left(
    \frac{1}{|Q_i|}\sum_{j\in Q_i}H_{\theta_0,j}^{L}(P_i^{\mathrm{text}})
    \right).
\end{equation}
For a visual-prompt template $P_i^{\mathrm{vis}}$, the prompt box $b_i$ and the
preprocessed image-token grid define a set $S_i$ of visual-token positions
inside the prompt region. We pool only those positions:
\begin{equation}
    z_i^{\mathrm{vis}}=\operatorname{normalize}\left(
    \frac{1}{|S_i|}\sum_{j\in S_i}H_{\theta,j}^{L}(P_i^{\mathrm{vis}})
    \right).
\end{equation}
PVRD-SG minimizes cosine distance between the prompt-region representation and
the cached text target:
\begin{equation}
    \mathcal{L}_{\sgloss}(\theta)=
    \E_{i\in\I_{\mathrm{vis}}}\left[1-(z_i^{\mathrm{vis}})^\top\sg(t_i)\right].
    \label{eq:sgloss}
\end{equation}
By construction, the text target contains $q_i$ but neither $a_i$ nor $r_i$.
Freezing $\theta_0$, caching $t_i$, and stopping its gradient keep the target fixed.
The readout excludes both the visual evidence outside $b_i$ and assistant
target positions. The loss is read from selected positions, but shared
parameters can still propagate its effect beyond the box. PVRD-SG therefore
does not teach the answer through its target. It teaches the visual prompt
region where the task semantics represented by the typed question should be
available, while the ordinary autoregressive loss remains responsible for the
reasoning trace and final answer.

\subsection{PRMLP: Masked latent prediction}

PVRD-SG specifies a text-side semantic target, but it does not directly require
the visual representation to remain stable when parts of the rendered question
are difficult to observe. PRMLP supplies this image-side constraint. It is a
latent consistency objective, not pixel reconstruction or transcript
generation: the representation read from a partially masked prompt region must
approach the representation of the corresponding clean crop. On a scheduled
PRMLP update, $\mathcal{M}_{\rho,\omega}$ masks a fraction $\rho$ of the prompt
region under block-sampling rule $\omega$, while leaving the rest of the image
unchanged. Let $x_i^{\mathrm{vis}}$ be the VTS composite or real-world image:
\begin{equation}
    x_i^{M}=\mathcal{M}_{\rho,\omega}(x_i^{\mathrm{vis}};b_i).
\end{equation}
The online branch receives $x_i^{M}$, and the target branch receives
the clean prompt crop $c_i$:
\begin{align}
    z_i^{\mathrm{full}} &= \pool\left(H_\theta(q_{\min},x_i^{M}),S_i\right), \\
    z_i^{\mathrm{crop}} &= \pool\left(H_\theta(q_{\min},c_i),S_i^{c}\right).
\end{align}
Here $S_i$ selects the prompt-region visual tokens in the masked composite, and
$S_i^c$ contains the visual tokens in the crop view. The token mask is mapped
from the stored box after image preprocessing. The loss is
\begin{equation}
    \mathcal{L}_{\prmlp}(\theta)=
    \E_{i\in\I_{\mathrm{vis}}}\left[
    1-\cos\left(q_\phi(z_i^{\mathrm{full}}),\sg(z_i^{\mathrm{crop}})\right)
    \right].
    \label{eq:prmlp}
\end{equation}
Both branches use the current model, and the clean-crop branch is detached.
$q_\phi$ is an optional predictor; the reported implementation sets it to the
identity. The target therefore changes as $\theta$ changes; there is no
exponential-moving-average encoder. PRMLP is applied to visual-prompt examples
and can run every $k$ steps to limit its additional cost. It uses no
transcript supervision. Because the target is visual, PRMLP can also be
applied to real-world prompts whose typography and layout differ from the VTS
renderer, provided that the training example supplies a prompt region and its
clean crop.

The complete supervised recipe is
\begin{equation}
    \mathcal{L}_{\mathrm{total}} = \mathcal{L}_{\sft}
    + \lambda_{\sgloss}\mathcal{L}_{\sgloss}
    + \lambda_{\prmlp}\mathcal{L}_{\prmlp},
\end{equation}
where both loss weights are fixed before training and recorded in the run
manifest.

The objectives supervise different relationships. PVRD-SG specifies which
typed task semantics should be recoverable from the prompt region; PRMLP
preserves the region's visual representation under partial occlusion. Their
shared answer target and common prompt-region readout keep both signals tied to
the same task-bearing pixels. Neither objective changes the inference
interface, and neither supplies an answer or a transcript at test time.

\subsection{GSPO continuation}

We continue the model trained with both PVRD-SG and PRMLP using GSPO on a
separate mixture of original and VTS examples. The raw reward is
\begin{equation}
    R(\hat{y},a)=0.1R_{\mathrm{format}}(\hat{y})
    +0.9R_{\mathrm{ans}}(\hat{y},a).
\end{equation}
Both terms are binary. $R_{\mathrm{format}}$ checks the output grammar, and
$R_{\mathrm{ans}}$ applies a deterministic verifier to the parsed final answer.
The coefficients specify the raw reward, not each term's share of the policy
gradient, and the reward provides no separate prompt-transcription supervision.
We refer to the resulting checkpoint as the GSPO continuation.

\section{Experiments and analysis}
\label{sec:experiments}

The experiments ask two questions: how much does reasoning change when a
question becomes pixels, and how much of that loss can prompt-region grounding
recover? The primary evaluation uses MATH-Vision, MathVista, ChartQA, and MMMU
\citep{wang2024mathvision,lu2024mathvista,masry2022chartqa,yue2024mmmu}.
Unless a table states otherwise, trained conditions report mean accuracy over
completed runs. ``Mean'' is the unweighted average of these four tasks.
Accuracy is reported in percent; gains and gaps are percentage points.

\subsection{Experimental setup}

The Original view keeps the question in the text channel. VTS renders the same
question into the image and supplies only a short fixed cue. The visual evidence
and answer remain paired. We report Original accuracy, VTS accuracy, and their
gap, $\Delta=\operatorname{Acc}^{\orig}-\operatorname{Acc}^{\vts}$.

Our main adaptation experiments use Qwen3-VL-4B-Instruct
\citep{bai2025qwen3vl}. The common supervised pool contains 24,761 controlled
source examples, each exported once as Original and once as VTS, plus 50,389
real-world visual-prompt examples. This produces 99,911 training views from
75,150 source examples. A separate set of 1,000 real-world examples is held out
before export and is never used for SFT or GSPO. Every adaptation condition
uses the same training export, assistant-target policy, initialization,
optimizer, checkpoint rule, and evaluation protocol.

Balanced Replay and SFT are distinct controls rather than consecutive stages.
Balanced Replay follows the standard supervised schedule and measures what the
mixed Original/VTS exposure provides by itself. SFT uses the same interface
mixture and ordinary next-token objective but continues until its measured
training cost matches Ours. The cost ledger counts all forward, target, and
backward passes introduced by PVRD-SG and PRMLP; SFT is stopped when its
training FLOPs are within 5\% of Ours. GPU-hours are recorded as a secondary
system measure but are not the matching criterion. Comparing Ours with
Balanced Replay tests the gain over the standard mixed-interface recipe, while
comparing it with SFT tests whether ordinary additional optimization is
sufficient. We use the latter as the primary baseline for method gains. Full
data, hyperparameters, and evaluator details are in the supplement.

\subsection{The semantic channel gap}

Accuracy falls under VTS for every evaluated model and task. We compare
thinking and instruct variants of Qwen3-VL
\citep{bai2025qwen3vl}, InternVL3.5 \citep{wang2025internvl35}, and the
reasoning model DeepEyes \citep{zheng2025deepeyes}. The complete task-level
table is in the supplement. For both Qwen3-VL sizes, the thinking variants are
stronger on typed questions but lose more when the question moves into the
image. On MATH-Vision, Qwen3-VL-4B-Thinking scores 60.0 under Original,
compared with 51.6 for Qwen3-VL-4B-Instruct, yet its gap is 27.4 rather than
16.8 points.

A small gap is not sufficient when both accuracies are low. DeepEyes-7B has
the smallest average gap but reaches only 26.6 Original accuracy on
MATH-Vision. Figure~\ref{fig:performance_findings} therefore keeps both
interface accuracies visible rather than ranking models by gap alone.
Construction controls support this interpretation. For Base, adding a blank
panel stays within 1.1 points of Original and duplicating the question across
both channels stays within 0.6, whereas removing the question causes accuracy
to collapse. Ours preserves this pattern and improves over Balanced Replay
across the tested renderer variants
(Appendix Table~\ref{tab:vts_validity_controls} and
Figure~\ref{fig:supp_robustness}).

\subsection{Grounding visual task semantics}

\begin{table*}[t]
\centering
\small
\renewcommand{\arraystretch}{1.04}
\setlength{\tabcolsep}{0pt}
\begin{tabular*}{\textwidth}{@{\extracolsep{\fill}}lrrrrrrrr@{}}
\toprule
& \multicolumn{2}{c}{MATH-Vision} &
\multicolumn{2}{c}{MathVista} &
\multicolumn{2}{c}{ChartQA} &
\multicolumn{2}{c}{MMMU} \\
\cmidrule(lr){2-3}\cmidrule(lr){4-5}
\cmidrule(lr){6-7}\cmidrule(lr){8-9}
Method & Original & VTS & Original & VTS & Original & VTS & Original & VTS \\
\midrule
Balanced Replay & 52.0 & 45.6 & 73.9 & 68.4 & 85.0 & 74.4 & 68.3 & 57.2 \\
SFT & 51.5 & 44.3 & 73.9 & 65.1 &
83.7 & 72.3 & 67.4 & 50.1 \\
Ours & 52.8 & 52.1 & 74.5 & 72.1 & 85.3 & 78.5 & 68.6 & 62.5 \\
\bottomrule
\end{tabular*}
\caption{Four-task cost-matched adaptation. SFT follows the
ordinary supervised objective and is trained to the same measured cost as
Ours. Benchmarks form column groups, with Original and VTS views shown beneath
each benchmark.}
\label{tab:four_task_results}
\end{table*}

At the same measured training cost, Ours improves VTS accuracy over SFT on all
four tasks (Table~\ref{tab:four_task_results}). The gains are 7.8 points on
MATH-Vision, 7.0 on MathVista, 6.2 on ChartQA, and 12.4 on MMMU. Original
accuracy changes by 0.6 to 1.6 points. Averaged across tasks, VTS accuracy rises
from 58.0 to 66.3 and Original accuracy from 69.1 to 70.3, reducing the
interface gap from 11.2 to 4.0 points. The remaining gap is 0.7 points on
MATH-Vision and 2.4 on MathVista, compared with 6.8 on ChartQA and 6.1 on
MMMU.

The two baselines clarify where this improvement comes from. Balanced Replay
reaches 61.4 mean VTS accuracy and 69.8 Original accuracy, so Ours adds 4.9
VTS points while retaining the typed interface within 0.5 point. Cost-matched
SFT reaches only 58.0 VTS accuracy despite receiving the larger ordinary
supervised budget. Ours exceeds it by 8.4 points on VTS and 1.2 points on
Original. The result does not imply that longer SFT must always underperform a
shorter schedule. It shows that, under the matched data, optimizer, checkpoint,
and cost protocol used here, additional next-token training does not account
for the gain from the region-level objectives.

\begin{figure}[t]
\centering
\includegraphics[width=\columnwidth]{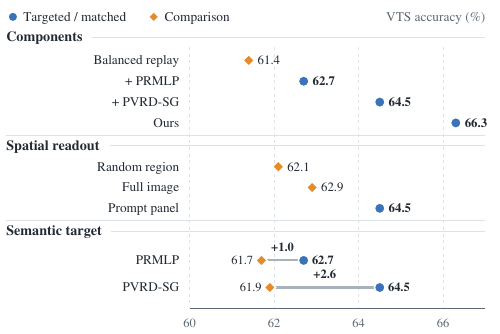}
\caption{Component contributions and specificity. Marks show four-task
mean VTS accuracy. The middle group compares spatial readouts at matched
training cost; the bottom group contrasts matched and deranged targets.
Original accuracy ranges from 69.8 to 70.3 across the displayed conditions.}
\label{fig:components_specificity}
\end{figure}

PVRD-SG accounts for the larger single-component gain, while PRMLP adds 1.8
VTS points over PVRD-SG when the objectives are combined
(Figure~\ref{fig:components_specificity}). Original accuracy varies by at most
0.5 points across these rows. Reading from the known prompt panel outperforms
full-image and random-region summaries at matched training cost. Matching each
target to its own example also beats a deranged target by 2.6 points for
PVRD-SG and 1.0 for PRMLP. The benefit therefore depends on both where the
representation is read and which semantic target it receives.
The representation probes follow the objective design: PVRD-SG primarily
improves typed-question retrieval, while PRMLP primarily improves clean-crop
retrieval (Appendix Table~\ref{tab:supp_representation_checks}).

The advantage also grows throughout training rather than appearing at one
selected checkpoint. At 25\%, 50\%, 75\%, and 100\% of the matched budget,
Ours exceeds Balanced Replay by 1.2, 2.5, 4.1, and 4.9 VTS points,
respectively, while Original accuracy stays within 0.5 points
(Appendix Figure~\ref{fig:supp_robustness}).

\subsection{Recognition is not task execution}

\begin{table}[t]
\centering
\small
\setlength{\tabcolsep}{1.2pt}
\begin{tabular*}{0.95\columnwidth}{@{\extracolsep{\fill}}lrrrrrrr@{}}
\toprule
Checkpoint & EM & Direct & Self & GT & P-crop & S-crop & S-gain \\
\midrule
Base & 87.6 & 48.4 & 56.1 & 58.3 & 54.6 & 49.3 & $+7.7$ \\
Balanced Replay & 92.4 & 57.0 & 60.2 & 61.5 & 59.7 & 57.8 & $+3.2$ \\
Ours & 95.1 & 62.1 & 62.9 & 63.4 & 63.0 & 62.2 & $+0.8$ \\
\bottomrule
\end{tabular*}
\caption{Same-composite recognition and use diagnostic. Two-task
means on the frozen composite. Self/GT reinsert transcripts; P-crop/S-crop add
token-matched prompt/scene crops; EM is exact match transcription.}
\label{tab:same_composite}
\end{table}

All answer conditions retain the same frozen VTS composite. \emph{Direct}
asks the model to answer that composite with the fixed cue used throughout VTS.
\emph{Self} adds the same checkpoint's cached transcription to the text channel,
whereas \emph{GT} adds the ground-truth question. Thus Self minus Direct
measures how much answer accuracy is recovered when the model's recognized
words regain the native text-channel role, and GT minus Self estimates the
remaining effect of transcription errors. \emph{P-crop} adds a clean crop of
the prompt region as a second image. \emph{S-crop} adds a scene crop sampled
outside that region with matched dimensions, aspect ratio, image order,
visual-token count, and auxiliary cue. Their difference tests whether isolating
the task-bearing region helps beyond supplying another image crop. EM is exact
match between the cached transcription and the ground-truth question.

The base model already transcribes the prompt well, yet reinserting its own
transcript improves answer accuracy by 7.7 points
(Table~\ref{tab:same_composite}). After full adaptation, Direct accuracy is
13.7 points higher and the Self advantage falls to 0.8; replacing Self with
ground-truth text adds only another 0.5 point. Thus recognition alone does not
explain the base failure, while adaptation reduces the benefit of moving the
recognized question back into the text channel.

The crop interventions lead to the same interpretation from the visual side.
For Base, the prompt crop improves Direct accuracy by 6.2 points, while the
matched scene crop adds only 0.9. Ours answers the full composite at 62.1 and
reaches 63.0 with the prompt crop and 62.2 with the scene crop. The prompt
region is therefore useful when isolated for the base model, but after
adaptation the full composite already provides nearly all of that benefit.

Prompt-swap and fixed-OCR interventions show the same reduction in reliance on
text-channel reinsertion; their complete results are in Appendix
Table~\ref{tab:supp_behavioral_controls} and
Figure~\ref{fig:supp_diagnostics}.

\subsection{RL refinement}

\begin{table}[t]
\centering
\small
\setlength{\tabcolsep}{3.6pt}
\begin{tabular*}{0.95\columnwidth}{@{\extracolsep{\fill}}lrrrr@{}}
\toprule
Checkpoint & Original & VTS & Gap & VTS gain \\
\midrule
Continued SFT & 70.4 & 67.8 & 2.6 & baseline \\
GSPO & 70.7 & 69.2 & 1.5 & $+1.4$ \\
\bottomrule
\end{tabular*}
\caption{Cost-matched GSPO continuation on Qwen3-VL-8B-Instruct.
Values are four-task mean accuracies.}
\label{tab:rlvr_main}
\end{table}

At matched training cost, GSPO improves mean VTS accuracy by 1.4 points over
continued SFT without reducing Original accuracy
(Table~\ref{tab:rlvr_main}); format accuracy also rises from 98.0 to 99.3.
The cost-matched InternVL3.5-8B transfer result is reported in Appendix
Table~\ref{tab:supp_internvl_task_level}.

\subsection{Independent benchmark transfer}

\begin{table}[t]
\centering
\small
\setlength{\tabcolsep}{3.8pt}
\begin{tabular*}{0.95\columnwidth}{@{\extracolsep{\fill}}lrrr@{}}
\toprule
\textit{VISTA-Bench} & VT & Text & VT gain \\
\midrule
Base & 52.3 & 59.1 & -- \\
Balanced Replay & 55.4 & 59.3 & -- \\
SFT & 56.2 & 59.4 & baseline \\
Ours & 60.3 & 59.6 & $+4.1$ \\
Ours $+$ GSPO & 60.1 & 61.3 & $+3.9$ \\
\midrule
\textit{OCRBench v2} & ZH & EN & Avg gain \\
\midrule
Base & 61.2 & 65.4 & -- \\
Balanced Replay & 59.7 & 65.2 & -- \\
SFT & 59.1 & 62.8 & baseline \\
Ours & 60.5 & 65.9 & $+2.3$ \\
Ours $+$ GSPO & 60.7 & 65.3 & $+2.1$ \\
\bottomrule
\end{tabular*}
\caption{External evaluations with distinct protocols. VISTA reports
official weighted VT/Text accuracy and VT gains over SFT; its Base row is
Qwen3-VL-8B-Instruct. OCRBench v2 reports the official ZH and EN scores; avg
gain is the unrounded mean of the language-specific gains over SFT.}
\label{tab:external_results}
\end{table}

On the independently constructed VISTA-Bench pairs, Ours raises weighted VT
accuracy from 56.2 to 60.3 while Text accuracy changes from 59.4 to 59.6
(Table~\ref{tab:external_results}). With GSPO, VT reaches 60.1 and the
Text-to-VT gap is 1.2 points. On OCRBench v2, Ours scores 60.5/65.9 on ZH/EN,
a 2.3-point average gain over SFT; Ours $+$ GSPO gains 2.1 points.
OCRBench measures text-rich capability rather than semantic-channel
equivalence, so it is not pooled with VISTA. Category-level VISTA results and
other text-rich benchmarks are in the supplement.

\subsection{Real-world task images}

\begin{figure}[t]
\centering
\includegraphics[width=\columnwidth]{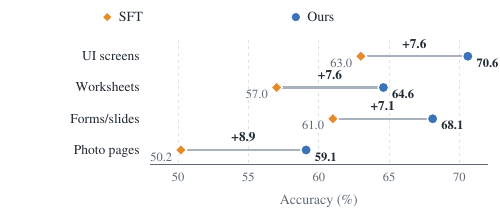}
\caption{Real-world natural-page evaluation. Accuracy under
SFT and Ours across four categories. Labels give the category scores and
gains over SFT.}
\label{fig:natural_transfer_main}
\end{figure}

We evaluate 1,000 real-world examples that are disjoint from training: 374 UI
screenshots, 188 worksheets, 157 forms or slides, and 281 photographed pages.
Ours improves over SFT in every category, by 7.1 to 8.9 points
(Figure~\ref{fig:natural_transfer_main}). Example-weighted accuracy rises from
58.0 to 65.8, a gain of 7.9 points; the unweighted category macro rises from
57.8 to 65.6. This set has no constructed typed counterpart, so it measures
generalization to the evaluated real-world pages rather than semantic-channel
equivalence.

\paragraph{Training controls.}
Question-scrubbed reasoning reaches 65.8 VTS accuracy, only 0.5 points below a
full trace that may repeat the question
(Appendix Table~\ref{tab:target_baseline_main}). The gain therefore does not
depend on copying the question into the assistant target. Ours also exceeds
VoQA QRA-SFT by 2.1 VTS points while retaining higher Original accuracy.

\section{Discussion and conclusion}
\label{sec:conclusion}

The same task is not equally usable across input channels. Moving a question
from text tokens into image pixels lowers accuracy in all 24 evaluated model
and task combinations, by 17.8 points on average. Blank-canvas and duplicate
question tests do not reproduce the loss, and the base model can transcribe
most visual questions exactly while still answering them poorly. The semantic
channel gap is therefore not well described as an OCR failure alone.

Prompt-region grounding raises four-benchmark VTS accuracy from 58.0 to 66.3
at matched cost while Original accuracy changes from 69.1 to 70.3.
Prompt-panel and deranged-target controls tie the gain to the intended spatial
and semantic signals without identifying a unique internal mechanism. Ours
also gains 4.1 points on the paired VISTA-Bench interface and 7.9 points on
held-out real-world pages; the supplement reports a cost-matched
second-backbone test. Inference remains one direct model call without an OCR
transcript or prompt box.

\bibliography{references}

\appendix
\raggedbottom
\renewcommand{\topfraction}{0.95}
\renewcommand{\bottomfraction}{0.85}
\renewcommand{\textfraction}{0.05}
\renewcommand{\floatpagefraction}{0.75}
\renewcommand{\dbltopfraction}{0.95}
\renewcommand{\dblfloatpagefraction}{0.75}
\setcounter{topnumber}{4}
\setcounter{bottomnumber}{2}
\setcounter{totalnumber}{6}
\setcounter{dbltopnumber}{4}
\setlength{\textfloatsep}{8pt plus 2pt minus 2pt}
\setlength{\floatsep}{8pt plus 2pt minus 2pt}
\setlength{\intextsep}{8pt plus 2pt minus 2pt}
\setlength{\dbltextfloatsep}{8pt plus 2pt minus 2pt}
\setlength{\dblfloatsep}{8pt plus 2pt minus 2pt}

\clearpage

\section{Supplementary overview}

The supplement follows the paper's main argument. It first clarifies the
relationship to prior visual-question work and gives the implementation of
prompt-region grounding. It then reports the full semantic channel gap, data
construction, training details, additional evaluations, and complete
qualitative rollouts. The final section states the limits of the evidence and
the release considerations for real-world images.

\section{Relationship to prior visual-question work}
\label{app:additional_related_work}

\paragraph{Visual questions and visualized text.}
VIM studies visual instructions and cross-interface mixture training
\citep{li2024textimages}. VoQA places the scene and question in one image and
reconstructs the visual question during supervised fine-tuning
\citep{an2025voqa}. VISTA-Bench evaluates matched text and visualized-text
questions under multiple rendering and OCR conditions
\citep{liu2026vistabench}. These studies establish that visualized questions
can be difficult. VTS complements them with a source-fixed intervention,
construction controls, and prompt-region supervision. We do not claim that
moving a question into pixels is itself new.

\paragraph{Local alignment and reasoning post-training.}
Region--text alignment and teacher--student vision-to-text distillation predate
our objectives
\citep{zhong2021regionclip,feng2025alignkd,chen2026alignti}. Our method uses the
known task-bearing region of a rendered training example as the readout for two
representation-level targets. For the final stage, GSPO clips importance ratios
at the response-sequence level \citep{zheng2025gspo}; we apply it with
verifiable answer and format rewards after supervised grounding.

\section{Implementation of prompt-region grounding}
\label{app:prompt_region_grounding_details}

Both auxiliary objectives use a prompt-region box and clean crop. The renderer
stores them for controlled VTS rows, while the GLM-OCR pipeline extracts them for real-world
rows. After image preprocessing, the box is mapped to the corresponding
visual-token region and the selected states are pooled. PVRD-SG aligns this
prompt-region representation with a cached embedding of the typed question as
a stop-gradient target. PRMLP masks random blocks inside the prompt region and
predicts the detached representation of the clean crop. The reported formal
configuration uses an identity predictor. Neither objective asks the model to
transcribe the prompt.

Capturing only the final normalized visual states avoids materializing every
decoder layer. PVRD-SG reuses the visual-prompt supervised forward pass. PRMLP
adds a masked-composite view and a clean-crop target view on its scheduled
updates. Prompt boxes, cached targets, and crop views are used only during
training; evaluation uses one composite image and the fixed text cue.

\section{Full semantic channel gap results}
\label{app:complete_vts_diagnostic}

\begin{table*}[t]
\centering
\small
\renewcommand{\arraystretch}{1.04}
\setlength{\tabcolsep}{0pt}
\begin{tabular*}{\textwidth}{@{\extracolsep{\fill}}l*{9}{r}@{}}
\toprule
& \multicolumn{2}{c}{MATH-Vision} &
\multicolumn{2}{c}{MathVista} &
\multicolumn{2}{c}{ChartQA} &
\multicolumn{2}{c}{MMMU} &
\shortstack{Mean\\gap} \\
\cmidrule(lr){2-3}\cmidrule(lr){4-5}
\cmidrule(lr){6-7}\cmidrule(lr){8-9}
Model & Original & VTS & Original & VTS & Original & VTS & Original & VTS & \\
\midrule
Qwen3-VL-4B-T & 60.0 & 32.6 & 79.5 & 56.4 & 88.8 & 60.8 & 70.8 & 43.2 & 26.5 \\
Qwen3-VL-4B-I & 51.6 & 34.8 & 73.7 & 61.9 & 84.6 & 67.8 & 67.4 & 49.6 & 15.8 \\
Qwen3-VL-8B-T & 62.7 & 35.7 & 81.4 & 60.8 & 88.6 & 63.4 & 74.1 & 46.5 & 25.1 \\
Qwen3-VL-8B-I & 53.9 & 38.6 & 77.2 & 66.0 & 89.6 & 73.2 & 69.6 & 54.1 & 14.6 \\
InternVL3.5-8B-I & 56.8 & 39.4 & 78.4 & 66.7 & 86.7 & 70.6 & 73.4 & 55.5 & 15.8 \\
DeepEyes-7B & 26.6 & 18.4 & 70.1 & 61.9 & 78.5 & 67.2 & 58.6 & 49.7 & 9.2 \\
\bottomrule
\end{tabular*}
\caption{\textbf{Original/VTS accuracy for all 24 model--task pairs.}
Each benchmark separates Original and VTS accuracy. T and I denote thinking
and instruct variants; Mean gap is the unweighted four-task mean of Original
minus VTS.}
\label{tab:all_model_gap_supp}
\end{table*}

Accuracy decreases in all 24 combinations. The thinking variants obtain
stronger Original scores than their instruct counterparts on several tasks,
but they also have larger mean gaps. DeepEyes has the smallest mean gap and
the lowest Original MATH-Vision accuracy, which is why we report both
interface accuracies rather than gap alone.

\section{Data, training, and evaluation}
\label{app:training_details}

\paragraph{Supervised grounding data.}
The supervised pool contains 75,150 source examples. Its controlled portion
has 24,761 examples, each exported in paired Original and VTS views, yielding
49,522 training views. VTS views store the full composite, prompt-panel
coordinates, typed-question target representation, and clean prompt crop;
Original views retain the native question and source image. The other 50,389
source examples are held-in real-world visual prompts described below. Each
retains its original text prompt alongside the task-bearing image. The GLM-OCR pipeline
provides a prompt box and clean crop for each real-world example. The resulting
training export has 99,911 views. All conditions use this same export
and assistant-target policy. Ours applies PVRD-SG and PRMLP to the 24,761 VTS
views and 50,389 real-world examples; there is no separate real-world SFT
stage.

The controlled examples are sampled from MMR1~\citep{leng2025mmr1},
BMMR~\citep{xi2025bmmr}, Euclid30K~\citep{lian2025euclid},
MMK12~\citep{meng2025mmeureka}, FineVision
subsets~\citep{wiedmann2025finevision}, mmopenr1-8k
\citep{lin2026mmfinereason}, and WeMath2 subsets
\citep{qiao2025wemath2}.

\begin{table}[t]
\centering
\small
\begin{tabular*}{0.95\columnwidth}{@{\extracolsep{\fill}}lr@{}}
\toprule
Data block & Source examples \\
\midrule
Rendered training split & 24,761 \\
Real-world training split & 50,389 \\
\midrule
Total supervised pool & 75,150 \\
\bottomrule
\end{tabular*}
\caption{\textbf{Source examples in the common supervised grounding pool.}
Each controlled example produces one Original and one VTS training view, so
24,761 controlled examples yield 49,522 views. Together with the real-world
split, the optimizer sees 99,911 views. The 1,000-example real-world validation
split is excluded from every training condition.}
\label{tab:training_data_sources}
\end{table}

\subsection{Real-world data and annotation}
\label{app:real_world_collection}

\paragraph{Collection and split.}
We collected 51,389 real-world image and question pairs. Before training, we set
aside 1,000 examples for the natural-page evaluation and removed them from the
SFT and GSPO exports. The remaining 50,389 examples are merged with the
24,761 controlled source examples in
Table~\ref{tab:training_data_sources}, yielding 75,150 source examples. After
the controlled examples are expanded into paired views, the supervised export
contains 99,911 training views. Each real-world example retains its original
text prompt as the semantic target and receives both region-level objectives
in addition to the autoregressive supervised loss.

\paragraph{Annotation workflow.}
The GLM-OCR pipeline~\citep{duan2026glmocr} first extracts the task text, its bounding box,
and the corresponding crop from each image. GPT-5.4 then reads the source
image and extracted prompt, annotates the question, and produces a reasoning
trace with a final answer. Gemini 2.5 Pro performs the final quality check by
testing whether the reasoning supports the answer. Only examples that pass
this check are exported for training.
Independent college-student review agreed with the finalized annotations on
97.4\% of a 1,500-example sample.

\paragraph{GSPO data selection.}
We mine the GSPO split from the finalized SFT pool. For each source problem,
the supervised checkpoint produces eight independent completions, which are
scored by the deterministic answer verifier. Let
$c_i=\sum_{j=1}^{8}\mathbf{1}[\hat{a}_{ij}=a_i]$ be the number of correct
completions. We retain examples with $1\leq c_i\leq5$. This pass@8-style filter
removes unsolved cases with no positive signal and nearly saturated cases with
six to eight correct completions. The retained 23,488 source problems are
exported in both Original and VTS form, giving 46,976 training rows. A separate
512-problem development split gives 1,024 paired rows. GSPO initializes from
the Ours supervised checkpoint and uses
$0.1\,r_{\mathrm{format}}+0.9\,r_{\mathrm{answer}}$. The continued-SFT control
starts from the same checkpoint and uses matched training-cost accounting.

\begin{table*}[t]
\centering
\footnotesize
\renewcommand{\arraystretch}{1.02}
\begin{minipage}[t]{0.485\textwidth}
\vspace{0pt}
\centering
\textbf{Supervised grounding}\par\smallskip
\begin{tabular}{@{}p{0.27\linewidth}p{0.68\linewidth}@{}}
\toprule
Setting & Value \\
\midrule
Initialization & Qwen3-VL-4B-Instruct \\
Optimization data & 75,150 source examples: 24,761 controlled examples
become 49,522 paired views, plus 50,389 real-world views \\
Trainable modules & Full language model and multimodal projector; frozen
vision tower \\
Optimizer & AdamW; weight decay 0 \\
Learning rate & $5{\times}10^{-7}$; cosine decay; 3\% warmup \\
Batching & Global batch 8; one example per GPU; gradient accumulation 1 \\
Duration & 1 epoch; 12,489 optimizer steps \\
Text length & 8,192-token cutoff; PRMLP main/extra caps 4,096/2,048 \\
Image pixels & 1,024--12,845,056; PRMLP crop cap 262,144 \\
Precision & BF16 on 8 GPUs \\
PVRD-SG & $\lambda_{\mathrm{sg}}=0.15$ on VTS and real-world visual-prompt
rows; VTS language-model weight 1.0 \\
PRMLP & $\lambda_{\mathrm{prmlp}}=0.003$ on VTS and real-world visual-prompt
rows; 35\% block mask; 32-pixel blocks; update every 2 steps; detached current
target; identity predictor \\
Reward/target & Shared answer target for Original and VTS \\
\bottomrule
\end{tabular}
\end{minipage}
\hfill
\begin{minipage}[t]{0.485\textwidth}
\vspace{0pt}
\centering
\textbf{GSPO continuation}\par\smallskip
\begin{tabular}{@{}p{0.27\linewidth}p{0.68\linewidth}@{}}
\toprule
Setting & Value \\
\midrule
Initialization & Our supervised checkpoint \\
Optimization data & 46,976 rows: 23,488 Original and 23,488 VTS; 1,024
paired development rows \\
Trainable modules & Full actor under FSDP; frozen reference policy \\
Optimizer & AdamW-BF16; $(\beta_1,\beta_2)=(0.9,0.999)$; weight decay 0.01 \\
Learning rate & $1{\times}10^{-6}$; constant schedule; no warmup \\
Batching & Rollout and actor global batch 40; update and experience
microbatch 1 per GPU \\
Duration & 5 epochs; 5,870 actor-update steps \\
Text length & 4,096-token prompt; 12,288-token response; 16,384-token rollout
context \\
Image pixels & 1,024--589,824 \\
Precision & BF16 full-shard FSDP on $5{\times}8$ GPUs \\
Candidate mining & 8 completions per source; retain 1--5 correct completions \\
Policy sampling & 5 completions per prompt; temperature 1.0; top-$p=1.0$ \\
Clipping and KL & clip low/high 0.2/0.3; one policy epoch; max gradient norm
1.0; low-variance KL coefficient 0.01 \\
Reward & $0.1\,r_{\mathrm{format}}+0.9\,r_{\mathrm{answer}}$ \\
\bottomrule
\end{tabular}
\end{minipage}
\caption{\textbf{Training hyperparameters.} Candidate-mining completions
construct the GSPO training set; five policy completions per prompt are sampled
during optimization.}
\label{tab:training_hyperparameters}
\end{table*}

\begin{figure*}[t]
\centering
\includegraphics[width=\textwidth]{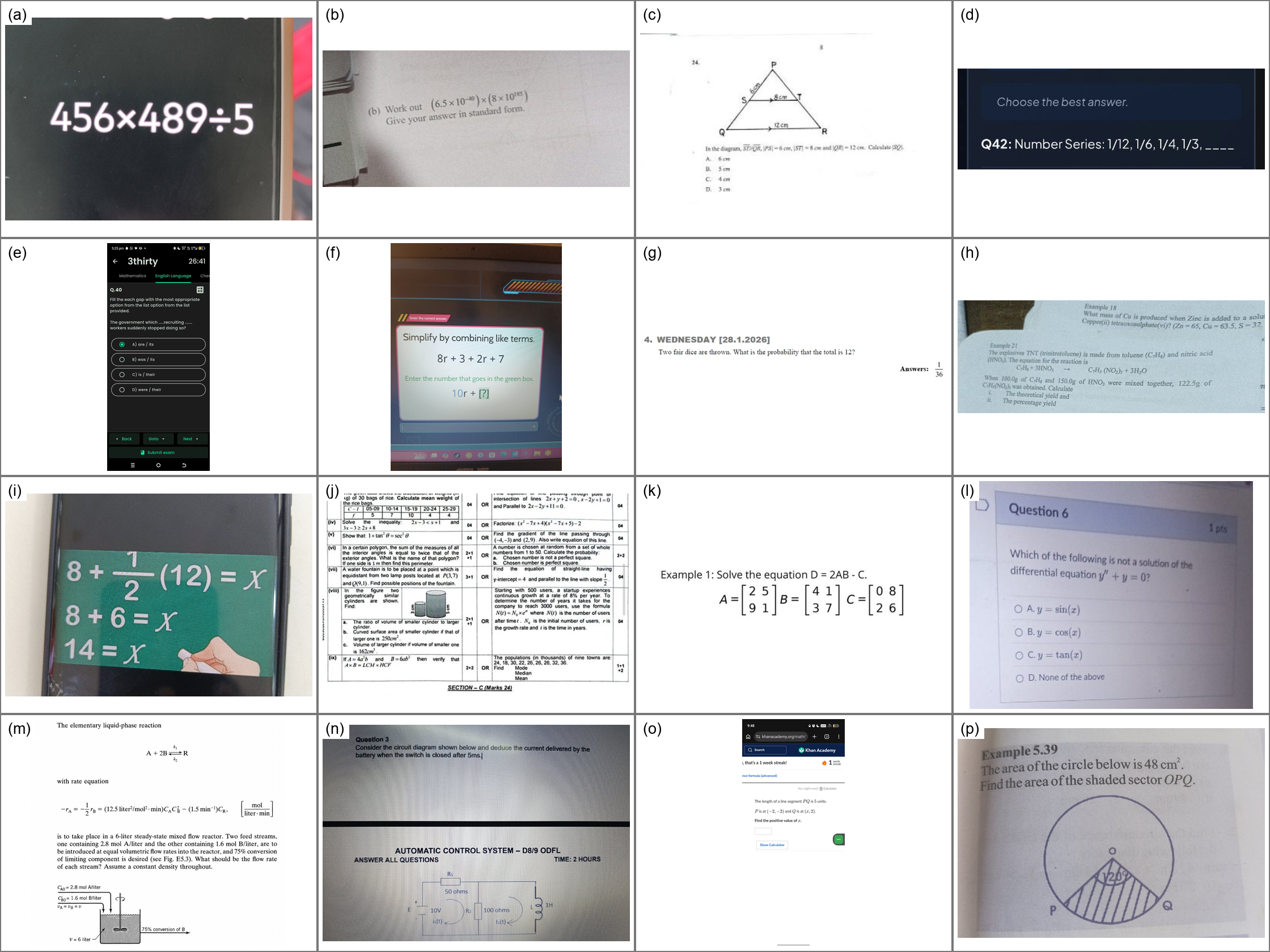}
\caption{\textbf{Representative samples from the real-world training
collection.} The examples span photographed screens and textbook pages,
worksheets and exam sheets, learning interfaces, slides, and other layouts in
which the task-defining text appears inside the visual input. These samples
come from the training split, not the held-out 1,000-example evaluation.}
\label{fig:training_set_examples}
\end{figure*}

\paragraph{Evaluation.}
MATH-Vision, MathVista, ChartQA, and MMMU use their standard answer extraction
and accuracy evaluators. Every method row reports Original and VTS on the same
task set. VISTA-Bench uses its official paired Text/VT examples and weighted
overall score. OCRBench v2, DocVQA, TextVQA, and ST-VQA retain their own
official metrics and are never averaged into a VTS headline. The natural-page
evaluation contains 1,000 real-world examples held out from training: 374 UI
screenshots, 188 worksheets, 157 forms/slides, and 281 photographed pages. It
uses each original visual input and reports standard answer accuracy; no paired
typed view is constructed.

\paragraph{Release.}
We plan to release the finalized annotations, split manifests, processing
scripts, and all images that pass the applicable consent, privacy, and license
checks.

\section{Further evidence}
\label{app:additional_reported_results}

\subsection{Prompt cues and supervision}

\begin{table}[t]
\centering
\scriptsize
\renewcommand{\arraystretch}{1.04}
\setlength{\tabcolsep}{0pt}
\begin{tabular*}{0.95\columnwidth}{@{\extracolsep{\fill}}lrrrr@{}}
\toprule
Checkpoint &
\shortstack{Original} &
\shortstack{Minimal} &
\shortstack{Native} &
\shortstack{Identify--solve} \\
\midrule
Base & 69.3 & 53.5 & 55.4 & 56.2 \\
Ours & 70.3 & 66.3 & 67.0 & 67.4 \\
\bottomrule
\end{tabular*}
\caption{\textbf{Cue robustness.} Values are four-task macro accuracies; cue
variants keep the VTS image fixed.}
\label{tab:supp_cue_target_controls}
\end{table}

\begin{table}[t]
\centering
\small
\setlength{\tabcolsep}{3.3pt}
\begin{tabular*}{0.95\columnwidth}{@{\extracolsep{\fill}}lrrr@{}}
\toprule
Training condition & Original & VTS & VTS gain \\
\midrule
\multicolumn{4}{@{}l}{\textit{Assistant target; gain over answer only}} \\
Answer only & 69.8 & 63.2 & baseline \\
Question-scrubbed trace & 70.1 & 65.8 & $+2.6$ \\
Full trace & 70.3 & 66.3 & $+3.1$ \\
\midrule
\multicolumn{4}{@{}l}{\textit{Prior recipes; gain over Balanced Replay}} \\
Balanced Replay & 69.8 & 61.4 & baseline \\
QA-SFT & 69.7 & 63.0 & $+1.6$ \\
VoQA QRA-SFT & 69.8 & 64.2 & $+2.8$ \\
\textbf{Ours} & \textbf{70.3} & \textbf{66.3} & \textbf{$+4.9$} \\
\bottomrule
\end{tabular*}
\caption{Training-target and closest-baseline comparisons. Values are
four-task mean accuracies. Target rows match source items, steps, and training
cost. Balanced Replay is the VIM-style mixed-interface baseline.}
\label{tab:target_baseline_main}
\end{table}

Changing the VTS cue improves the base-model accuracy from 53.5 to at most
56.2, whereas Ours reaches 66.3--67.4 under all three cues. The
channel gap therefore remains visible under stronger scaffolds. For assistant
targets, the question-scrubbed trace reaches 65.8 VTS accuracy, compared with
63.2 for answer-only and 66.3 for the full trace. Thus most of the trace benefit
does not require repeating the question. Ours also reaches 66.3 VTS
and 70.3 Original accuracy, compared with 64.2 and 69.8 for VoQA QRA-SFT.

\subsection{Training dynamics and prompt following}

The VTS gain grows from 1.2 points at one quarter of the budget to 4.9 points
at the completed budget (Figure~\ref{fig:supp_robustness}). Replay VTS
accuracy is 57.2/59.0/60.3/61.4 across the four cost fractions; Ours reaches
58.4/61.5/64.4/66.3, giving gains of 1.2/2.5/4.1/4.9. Original accuracy stays
within 0.5 points between the two recipes throughout the trajectory.

\begin{table}[t]
\centering
\scriptsize
\renewcommand{\arraystretch}{1.04}
\setlength{\tabcolsep}{0pt}
\begin{tabular*}{0.95\columnwidth}{@{\extracolsep{\fill}}lrrrrrr@{}}
\toprule
Checkpoint &
\shortstack{Direct} &
\shortstack{Swap} &
\shortstack{Wrong} &
\shortstack{Prompt} &
\shortstack{Evidence} &
Other \\
\midrule
Base & 53.5 & 25.1 & 13.2 & 31.2 & 52.4 & 16.4 \\
Balanced replay & 61.4 & 36.8 & 19.0 & 44.7 & 40.1 & 15.2 \\
Ours & 66.3 & 48.2 & 25.0 & 58.9 & 26.0 & 15.1 \\
\bottomrule
\end{tabular*}

\medskip

\begin{tabular*}{0.78\columnwidth}{@{\extracolsep{\fill}}lrrr@{}}
\toprule
Checkpoint &
\shortstack{Typed\\follow} &
\shortstack{Visual\\follow} &
Other \\
\midrule
Base & 80.5 & 8.1 & 11.4 \\
Balanced replay & 70.9 & 17.4 & 11.7 \\
Ours & 61.7 & 27.6 & 10.7 \\
\bottomrule
\end{tabular*}
\caption{\textbf{Prompt-swap and channel-conflict controls.} The
upper panel changes the visual prompt; the lower panel gives the two channels
different scorable targets.}
\label{tab:supp_behavioral_controls}
\end{table}

Prompt-follow accuracy rises from 31.2 for Base to 58.9 for Ours,
while evidence-follow accuracy falls from 52.4 to 26.0. Under direct channel
conflict, visual-instruction following rises by 10.2 points over balanced
replay. The latter result also motivates treating typed--visual instruction
priority as a separate safety question.

\subsection{Representation and PRMLP analyses}

\begin{table}[t]
\centering
\scriptsize
\renewcommand{\arraystretch}{1.04}
\setlength{\tabcolsep}{0pt}
\begin{tabular*}{0.95\columnwidth}{@{\extracolsep{\fill}}lrrrrr@{}}
\toprule
Training recipe &
VTS &
\shortstack{Text\\retr.} &
\shortstack{Crop\\retr.} &
\shortstack{Eff.\\rank} &
\shortstack{Off-diag.\\sim.} \\
\midrule
Balanced replay & 61.4 & 26.4 & 22.5 & 117.3 & 0.19 \\
Replay $+$ PVRD-SG & 64.2 & 48.7 & 24.0 & 113.7 & 0.21 \\
Replay $+$ PRMLP & 62.8 & 28.1 & 45.5 & 111.7 & 0.22 \\
Ours & 66.3 & 51.6 & 49.1 & 109.3 & 0.24 \\
\bottomrule
\end{tabular*}
\caption{\textbf{Representation sanity checks.} Retrieval uses a
fixed 512-item pool; rank and similarity help screen for collapse.}
\label{tab:supp_representation_checks}
\end{table}

PVRD-SG primarily raises text retrieval accuracy, whereas PRMLP primarily
raises crop retrieval accuracy. Ours improves both. Effective rank
remains above 109 and off-diagonal similarity remains at or below 0.24 across
the displayed recipes.

Figure~\ref{fig:supp_diagnostics} compares the PRMLP objective and schedule
under the same total-cost accounting. The formal masked target gives the
largest VTS gain over PVRD-SG only. The shorthand settings are: small
($\lambda=0.001$, every 4 steps), formal ($0.003$, every 2), frequent
($0.003$, every step), and high weight ($0.010$, every 2); the unmasked,
learned-predictor, and matched-token variants use $0.003$ every 2 steps.
Using a learned predictor or a matched-token crop remains close to the formal
setting, while the unmasked target and high-weight setting are weaker.
Original accuracy spans 69.6--70.3 across these configurations.

\begin{table*}[t]
\centering
\small
\renewcommand{\arraystretch}{1.04}
\setlength{\tabcolsep}{0pt}
\begin{tabular*}{\textwidth}{@{\extracolsep{\fill}}l*{10}{r}@{}}
\toprule
& \multicolumn{2}{c}{MATH-Vision} &
\multicolumn{2}{c}{MathVista} &
\multicolumn{2}{c}{ChartQA} &
\multicolumn{2}{c}{MMMU} &
\multicolumn{2}{c}{Mean} \\
\cmidrule(lr){2-3}\cmidrule(lr){4-5}
\cmidrule(lr){6-7}\cmidrule(lr){8-9}\cmidrule(lr){10-11}
Recipe & Original & VTS & Original & VTS & Original & VTS &
Original & VTS & Original & VTS \\
\midrule
Continued SFT & 52.9 & 54.0 & 74.6 & 73.5 & 85.4 & 79.7 &
68.7 & 63.9 & 70.4 & 67.8 \\
GSPO & 53.1 & 55.6 & 74.8 & 74.6 & 85.7 & 80.8 &
69.2 & 65.8 & 70.7 & 69.2 \\
\bottomrule
\end{tabular*}
\caption{\textbf{GSPO task-level continuation.} Original and VTS accuracy are
reported separately for every benchmark and as an unweighted four-task mean.
Both updates start from the same Ours checkpoint and use matched training
cost. Means use full-precision task scores before one-decimal display
rounding.}
\label{tab:supp_gspo_task_level}
\end{table*}

\begin{table*}[t]
\centering
\small
\renewcommand{\arraystretch}{1.04}
\setlength{\tabcolsep}{0pt}
\begin{tabular*}{\textwidth}{@{\extracolsep{\fill}}l*{10}{r}@{}}
\toprule
& \multicolumn{2}{c}{MATH-Vision} &
\multicolumn{2}{c}{MathVista} &
\multicolumn{2}{c}{ChartQA} &
\multicolumn{2}{c}{MMMU} &
\multicolumn{2}{c}{Mean} \\
\cmidrule(lr){2-3}\cmidrule(lr){4-5}
\cmidrule(lr){6-7}\cmidrule(lr){8-9}\cmidrule(lr){10-11}
Recipe & Original & VTS & Original & VTS & Original & VTS &
Original & VTS & Original & VTS \\
\midrule
Base & 56.8 & 39.4 & 78.4 & 66.7 & 86.7 & 70.6 &
73.4 & 55.5 & 73.8 & 58.1 \\
SFT & 57.0 & 47.8 & 78.5 & 73.0 & 86.8 & 78.0 &
73.7 & 64.0 & 74.0 & 65.7 \\
Ours & 57.2 & 53.2 & 78.7 & 75.5 & 87.0 & 81.3 &
73.9 & 69.4 & 74.2 & 69.8 \\
\bottomrule
\end{tabular*}
\caption{\textbf{InternVL3.5-8B task-level transfer.} Original and VTS
accuracy are reported separately under the same four evaluation protocols;
Mean is the unweighted four-task aggregate computed before one-decimal display
rounding.}
\label{tab:supp_internvl_task_level}
\end{table*}

\FloatBarrier

\subsection{Generalization beyond VTS}

\begin{table}[t]
\centering
\scriptsize
\renewcommand{\arraystretch}{1.04}
\setlength{\tabcolsep}{0pt}
\begin{tabular*}{0.95\columnwidth}{@{\extracolsep{\fill}}lrrr@{}}
\toprule
Benchmark & Baseline & Ours & Gain \\
\midrule
DocVQA & 68.2 & 69.5 & $+1.3$ \\
TextVQA & 71.8 & 72.4 & $+0.6$ \\
ST-VQA & 66.0 & 66.8 & $+0.8$ \\
\bottomrule
\end{tabular*}
\caption{\textbf{External text-rich benchmarks.} Baseline is cost-matched SFT,
and each benchmark retains its official score.}
\label{tab:supp_text_rich_transfer}
\end{table}

\begin{table}[t]
\centering
\scriptsize
\renewcommand{\arraystretch}{1.04}
\setlength{\tabcolsep}{0pt}
\begin{tabular*}{0.95\columnwidth}{@{\extracolsep{\fill}}lrrrr@{}}
\toprule
& \multicolumn{2}{c}{VIM} & \multicolumn{2}{c}{VoQA} \\
\cmidrule(lr){2-3}\cmidrule(lr){4-5}
Recipe & Text & Pixel & Traditional & Visual-only \\
\midrule
Base & 68.5 & 50.2 & 66.0 & 49.5 \\
Balanced replay & 68.7 & 59.8 & 66.2 & 58.9 \\
VoQA QRA-SFT & 68.1 & 61.0 & 66.4 & 62.8 \\
Ours & 69.0 & 63.5 & 66.8 & 64.0 \\
\bottomrule
\end{tabular*}
\caption{\textbf{Prior matched-interface protocols.} Each protocol retains
its own split, prompt, and evaluator; scores are not pooled.}
\label{tab:supp_prior_protocol_transfer}
\end{table}

Ours improves over SFT on DocVQA, TextVQA, and ST-VQA, with gains
from 0.6 to 1.3 points. These benchmarks measure text-rich capability rather
than a paired change in the channel carrying the task. On InternVL3.5-8B, Ours
improves VTS accuracy by 4.1 points over cost-matched SFT while Original
accuracy changes by 0.2 (Table~\ref{tab:supp_internvl_task_level}).
On prior matched-interface protocols, Ours is 3.7 points above balanced replay
on VIM Pixel and 5.1 points above it on VoQA Visual-only. Relative to VoQA
QRA-SFT, the gains are 2.5 and 1.2 points, respectively.

\begin{table}[t]
\centering
\scriptsize
\renewcommand{\arraystretch}{1.04}
\setlength{\tabcolsep}{0pt}
\begin{tabular*}{0.95\columnwidth}{@{\extracolsep{\fill}}lrrrrr@{}}
\toprule
Checkpoint & Clean & Priority & Benign & Visible & Obfus. \\
\midrule
Base & 53.5 & 93.8 & 21.0 & 11.2 & 6.8 \\
Balanced replay & 61.4 & 94.0 & 35.0 & 11.0 & 6.7 \\
Ours & 66.3 & 94.1 & 52.0 & 10.9 & 6.6 \\
\bottomrule
\end{tabular*}
\caption{\textbf{Visual-instruction safety check.} Columns report clean VTS
accuracy, higher-priority instruction accuracy, benign visual following, and
visible/obfuscated attack success.}
\label{tab:supp_visual_instruction_safety}
\end{table}

The full method raises benign visual-follow accuracy while higher-priority
instruction accuracy remains near 94\%. Visible and obfuscated attack success
do not increase across the displayed checkpoints. This check characterizes the
evaluated conflicts and is separate from the grounding analysis.

\begin{table*}[t]
\centering
\small
\renewcommand{\arraystretch}{1.04}
\setlength{\tabcolsep}{0pt}
\begin{tabular*}{\textwidth}{@{\extracolsep{\fill}}l*{10}{r}@{}}
\toprule
& \multicolumn{2}{c}{Multimodal perception} &
\multicolumn{2}{c}{Multimodal reasoning} &
\multicolumn{2}{c}{Multimodal knowledge} &
\multicolumn{2}{c}{Unimodal knowledge} &
\multicolumn{2}{c}{Weighted overall} \\
\cmidrule(lr){2-3}\cmidrule(lr){4-5}\cmidrule(lr){6-7}
\cmidrule(lr){8-9}\cmidrule(lr){10-11}
Condition & VT & Text & VT & Text & VT & Text & VT & Text & VT & Text \\
\midrule
Base & 65.3 & 67.3 & 49.0 & 49.3 & 37.8 & 48.5 & 58.2 & 68.4 & 52.3 & 59.1 \\
Balanced replay & 67.0 & 67.4 & 52.2 & 49.6 & 42.0 & 48.8 & 61.0 & 68.6 & 55.4 & 59.3 \\
SFT & 67.4 & 67.5 & 52.9 & 49.7 & 43.0 & 49.0 & 62.1 & 68.7 & 56.2 & 59.4 \\
Ours & 68.8 & 67.7 & 57.2 & 50.1 & 48.5 & 49.2 & 66.5 & 68.9 & 60.3 & 59.6 \\
\bottomrule
\end{tabular*}
\caption{\textbf{VISTA-Bench results.} Every category separates the official
VT and Text interfaces. Base is Qwen3-VL-8B-Instruct, and the overall accuracy
weights categories by their sample counts.}
\label{tab:supp_vista_categories}
\end{table*}

Relative to SFT, Ours improves VT accuracy by 1.4
points on multimodal perception, 4.3 on multimodal reasoning, 5.5 on
multimodal knowledge, and 4.4 on unimodal knowledge. Its weighted VT gain is
4.1 points, while weighted Text accuracy changes by 0.2 points.

\subsection{Renderer robustness and task use}

\begin{table}[t]
\centering
\small
\renewcommand{\arraystretch}{1.02}
\setlength{\tabcolsep}{2.5pt}
\begin{tabular*}{0.95\columnwidth}{@{\extracolsep{\fill}}lrrrr@{}}
\toprule
& \multicolumn{2}{c}{Base} & \multicolumn{2}{c}{Ours} \\
\cmidrule(lr){2-3}\cmidrule(lr){4-5}
Condition & MATH-V. & MathVista & MATH-V. & MathVista \\
\midrule
Original   & 51.6 & 73.7 & 52.8 & 74.5 \\
VTS        & 34.8 & 61.9 & 52.1 & 72.1 \\
Canvas     & 50.7 & 72.6 & 52.0 & 73.8 \\
Duplicate  & 52.0 & 74.3 & 53.2 & 75.0 \\
Image-only &  7.9 & 33.5 &  8.2 & 34.1 \\
\bottomrule
\end{tabular*}
\caption{VTS construction controls. Canvas keeps the typed question
and adds a blank panel; Duplicate presents it in both channels; Image-only
removes it from both.}
\label{tab:vts_validity_controls}
\end{table}

\begin{figure*}[t]
\centering
\includegraphics[width=\textwidth]{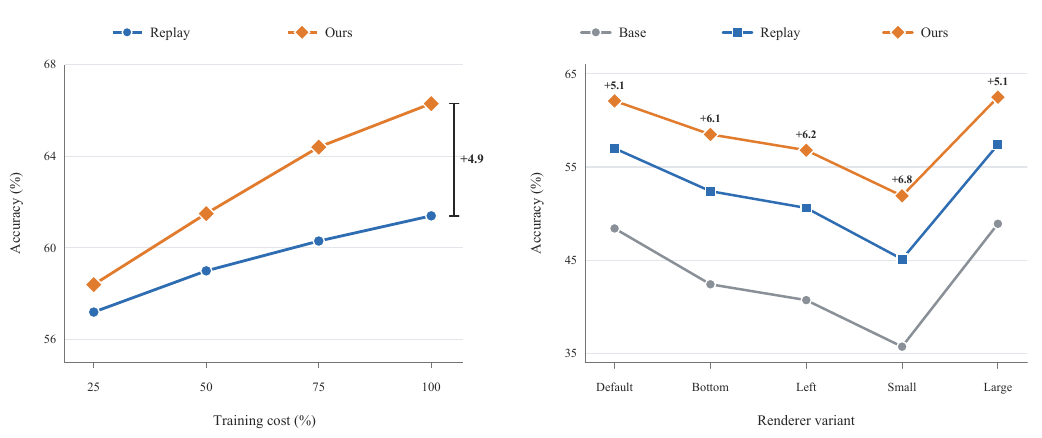}
\vspace{-0.45em}
\caption{\textbf{Training and renderer robustness.} Left: matched-budget VTS
learning curves at 25\%, 50\%, 75\%, and 100\% of the full training cost; the
final Ours-over-Replay difference is 4.9 points. Right: VTS accuracy when only
prompt-panel placement or font scale changes; labels give Ours-over-Replay
gains.}
\label{fig:supp_robustness}
\end{figure*}

Canvas stays within 1.1 points of Original for Base, and Duplicate stays
within 0.6. Image-only accuracy collapses because neither channel contains the
question (Table~\ref{tab:vts_validity_controls}).
Ours improves over balanced replay under every renderer. Smaller
text and unfamiliar placement remain harder than the default and large-font
conditions, so the result shows robustness within the tested renderer family
rather than invariance to layout (Figure~\ref{fig:supp_robustness}).

\begin{figure*}[t]
\centering
\includegraphics[width=\textwidth]{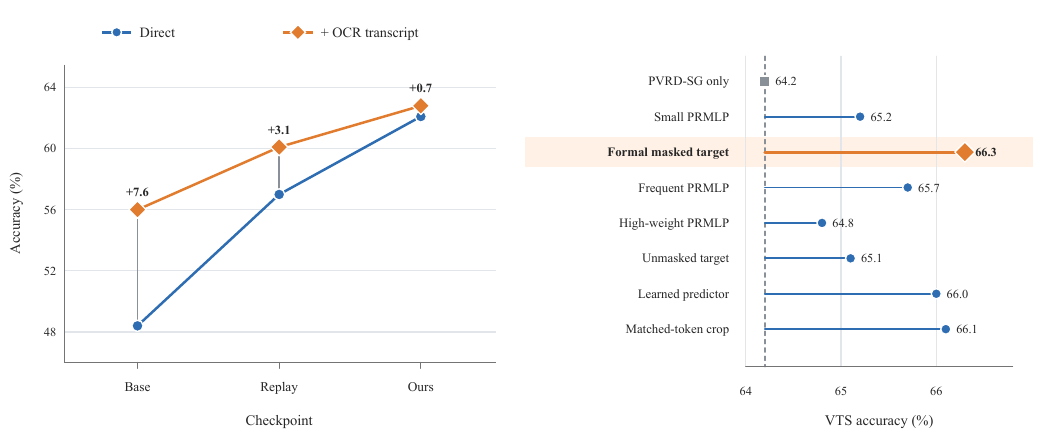}
\vspace{-0.45em}
\caption{\textbf{Task-use and PRMLP diagnostics.} Left: Direct and
OCR-transcript accuracy on the same VTS composite; the OCR gain contracts from
7.6 to 3.1 to 0.7 points. Right: four-task VTS accuracy under PRMLP variants;
the dashed line is PVRD-SG only, and the formal masked target is highlighted.}
\label{fig:supp_diagnostics}
\end{figure*}

For Base, supplying the OCR transcript on the unchanged composite adds 7.6
points, while restoring the source image adds a further 1.8 points. For the
full method, the corresponding gains are 0.7 and 0.2 points. The small
same-composite gain at the final supervised stage is consistent with answering
the composite directly (Figure~\ref{fig:supp_diagnostics}). With OCR fixed,
restoring the source adds 1.8, 0.5, and 0.2 points for Base, replay, and Ours;
ground-truth text adds a further 2.3, 1.4, and 0.6 points on the same composite.
On restored sources (Base/replay/Ours), fixed-cue, OCR, and ground-truth
scores are 20.7/21.0/21.2, 57.8/60.6/63.0, and 59.1/62.0/63.6.

\definecolor{casegood}{HTML}{2F6FB5}
\definecolor{casebad}{HTML}{B42318}
\newcommand{\caseerror}[1]{\textcolor{casebad}{#1}}
\newcommand{\rolloutformat}{%
  \footnotesize
  \setlength{\abovedisplayskip}{3pt}%
  \setlength{\belowdisplayskip}{3pt}%
  \setlength{\abovedisplayshortskip}{2pt}%
  \setlength{\belowdisplayshortskip}{2pt}%
}

\makeatletter
\setlength{\@dblfptop}{0pt}
\makeatother

\begin{figure*}[!t]
\section{Qualitative case studies}
\label{app:case_studies}
Aggregate accuracy hides how visual-task failures propagate. Each case below
pairs a complete, verified rollout from Our Model with an incorrect
Qwen3VL-8B-Thinking rollout. Red text marks the first recognition, reasoning,
or diagram error and every later quantity that depends on it. All four images
come from the held-out real-world split and were excluded from SFT and GSPO
training.

\vspace{0.45em}
\centering
\begin{minipage}[t]{0.31\textwidth}
\vspace{0pt}
\includegraphics[width=\linewidth,trim=0 25 0 315,clip]{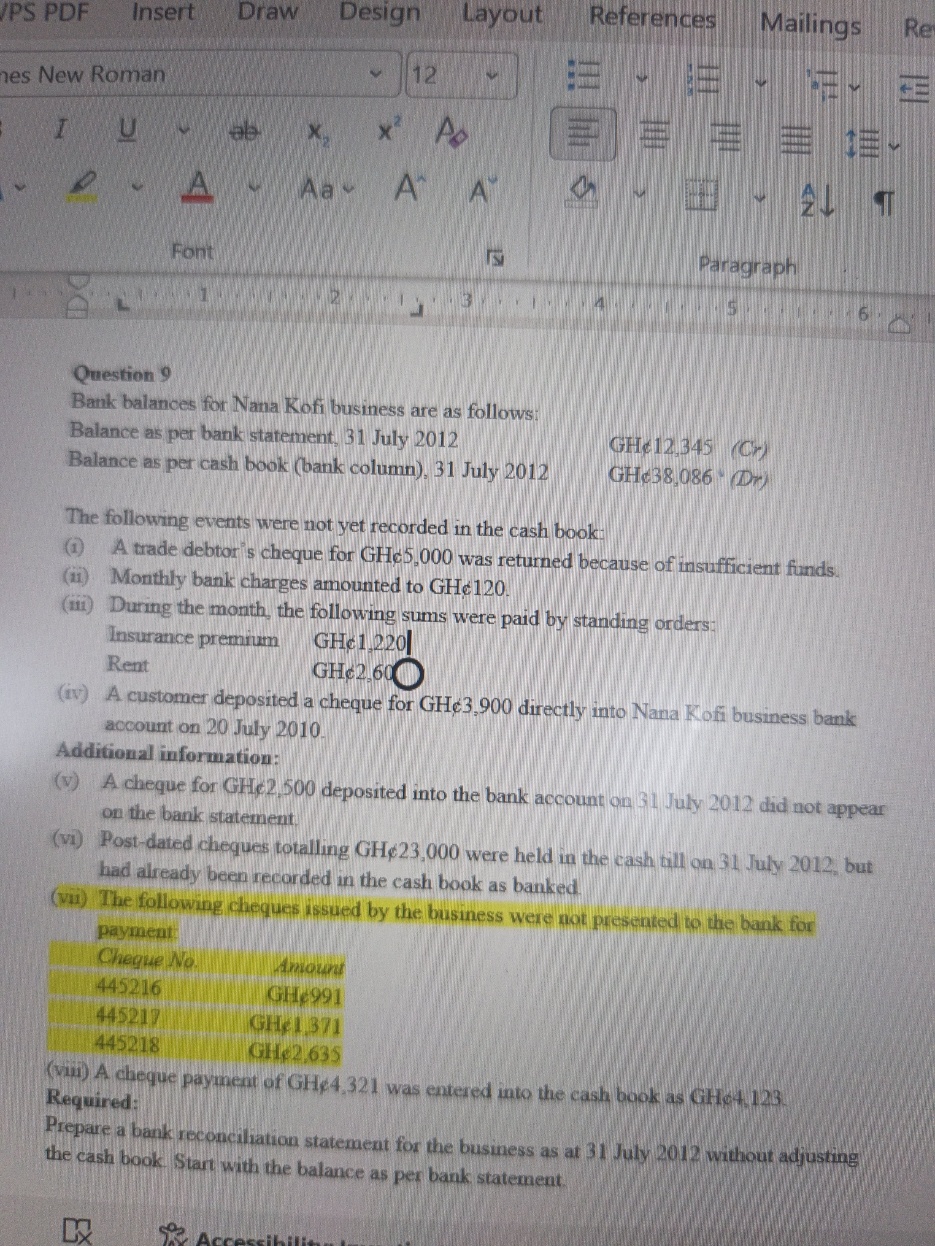}
\end{minipage}\hfill
\begin{minipage}[t]{0.655\textwidth}
\vspace{0pt}
\rolloutformat
\textbf{Bank reconciliation.}
The task asks for the unadjusted cash-book balance, starting from the
GH\textcent12,345 credit balance on the bank statement.

\smallskip
\textcolor{casegood}{\textbf{Our Model.}}
\textbf{1.} Add items that make the cash book higher than the statement:
the dishonoured cheque, bank charges, two standing orders, deposit in transit,
post-dated cheques recorded as banked, and the GH\textcent198 entry error.
\[
\Delta^{+}=5{,}000+120+1{,}220+2{,}600+2{,}500+23{,}000+198
=34{,}638.
\]
\textbf{2.} Subtract the direct deposit and the three unpresented cheques:
\[
991+1{,}371+2{,}635=4{,}997,\qquad
\Delta^{-}=3{,}900+4{,}997=8{,}897.
\]
\textbf{3.} Reconcile the balances:
\[
12{,}345+34{,}638-8{,}897
=\boxed{\text{GH\textcent}38{,}086\ \mathrm{Dr}}.
\]
This matches the cash-book balance printed in the image.

\smallskip
\textcolor{casebad}{\textbf{Qwen3VL-8B-Thinking.}}
\textbf{1.} The rollout transcribes
\(23{,}000\) as \(\caseerror{2{,}300}\) and
\(4{,}321\) as \(\caseerror{4{,}123}\).
\textbf{2.} The second error removes the required correction:
\[
\Delta_{\mathrm{entry}}
=\caseerror{4{,}123-4{,}123=0}.
\]
\textbf{3.} It keeps the other signs unchanged and computes
\[
\widetilde{\Delta}^{+}
=5{,}000+120+1{,}220+2{,}600+2{,}500
+\caseerror{2{,}300}+\caseerror{0}=13{,}740.
\]
\textbf{4.} It therefore returns
\[
12{,}345+13{,}740-8{,}897
=\boxed{\caseerror{\text{GH\textcent}17{,}188\ \mathrm{Dr}}}.
\]
The answer is short by GH\textcent20,898, exactly the omitted
GH\textcent20,700 and GH\textcent198.
\end{minipage}

\vspace{0.45em}
\hrule
\vspace{0.45em}

\begin{minipage}[t]{0.31\textwidth}
\vspace{0pt}
\includegraphics[width=\linewidth]{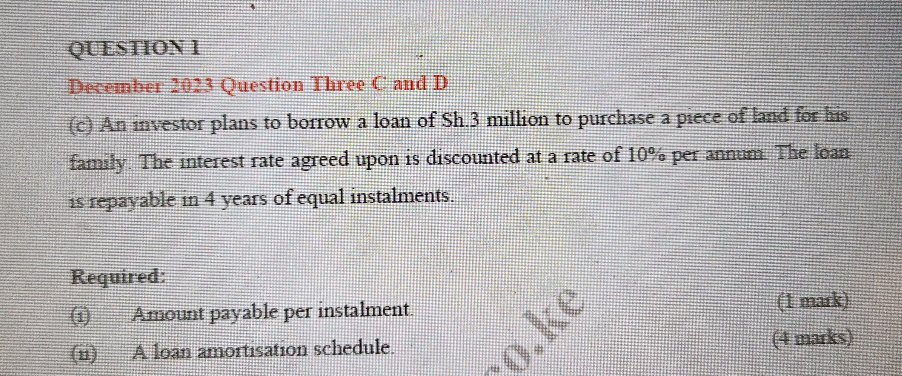}
\end{minipage}\hfill
\begin{minipage}[t]{0.655\textwidth}
\vspace{0pt}
\rolloutformat
\textbf{Equal-instalment loan.}
The loan principal is Sh 3,000,000, the annual rate is 10\%, and four equal
payments are made, one at the end of each year.

\smallskip
\textcolor{casegood}{\textbf{Our Model.}}
\textbf{1.} Equate the principal to the present value of an ordinary annuity:
\[
3{,}000{,}000
=A\frac{1-(1.10)^{-4}}{0.10}.
\]
\textbf{2.} Solve for the annual instalment:
\[
A=\frac{3{,}000{,}000(0.10)}{1-(1.10)^{-4}}
=\boxed{\text{Sh }946{,}412.41}.
\]
\textbf{3.} Let \(B_t\) be the balance after payment \(t\). Since each year's
interest is 10\% of its opening balance,
\[
B_t=1.10B_{t-1}-A,\qquad B_0=3{,}000{,}000.
\]
The schedule closes as
\[
\begin{aligned}
B_1&=2{,}353{,}587.59,& B_2&=1{,}642{,}533.94,\\
B_3&=860{,}374.92,& B_4&=0.00.
\end{aligned}
\]
The zero final balance verifies both the instalment and the amortization.

\smallskip
\textcolor{casebad}{\textbf{Qwen3VL-8B-Thinking.}}\par\noindent
\textbf{1.} The rollout treats the loan as carrying
\(\caseerror{\text{flat interest on the original principal}}\), so it charges
\[
I_t=\caseerror{0.10(3{,}000{,}000)=300{,}000}
\quad\text{in every year}.
\]
\textbf{2.} It sets total interest to
\(\caseerror{4(300{,}000)=1{,}200{,}000}\).
\textbf{3.} It divides principal plus flat interest by four:
\[
\widetilde A
=\frac{3{,}000{,}000+\caseerror{1{,}200{,}000}}{4}
=\boxed{\caseerror{\text{Sh }1{,}050{,}000}}.
\]
\textbf{4.} This cannot be the required equal instalment: its four discounted
payments have present value
\[
1{,}050{,}000\frac{1-(1.10)^{-4}}{0.10}
=\caseerror{\text{Sh }3{,}328{,}358.72}
\ne\text{Sh }3{,}000{,}000.
\]
\end{minipage}
\caption{\textbf{Finance rollouts.} Each example contrasts the verified
derivation from Our Model with a Qwen3VL-8B-Thinking rollout containing a
specific recognition or reasoning error.}
\label{fig:case_studies_finance}
\end{figure*}

\begin{figure*}[!t]
\centering
\begin{minipage}[t]{0.31\textwidth}
\vspace{0pt}
\includegraphics[width=\linewidth]{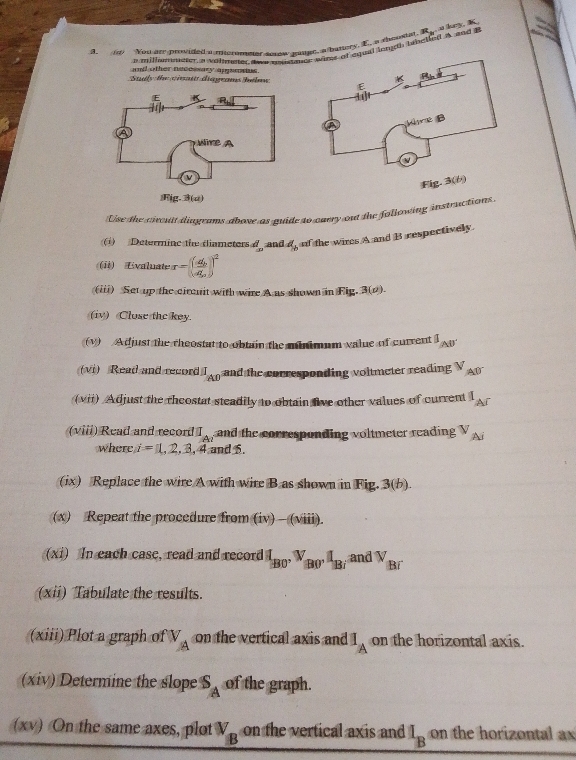}
\end{minipage}\hfill
\begin{minipage}[t]{0.655\textwidth}
\vspace{0pt}
\rolloutformat
\textbf{Two-wire circuit experiment.}
The procedure measures current and voltage for wires A and B, then plots
voltage on the vertical axis against current on the horizontal axis.

\smallskip
\textcolor{casegood}{\textbf{Our Model.}}
\textbf{1.} The ammeter is in series and the voltmeter is across the test
wire, so each pair \((I_X,V_X)\) refers to the same wire \(X\).
\textbf{2.} The requested axes give
\[
S_X=\frac{\Delta V_X}{\Delta I_X}=R_X.
\]
\textbf{3.} For wires of the same material and length,
\[
R_X=\frac{\rho L}{A_X}
=\frac{4\rho L}{\pi d_X^2}.
\]
\textbf{4.} Therefore,
\[
\boxed{\frac{S_A}{S_B}
=\frac{R_A}{R_B}
=\left(\frac{d_B}{d_A}\right)^2}.
\]
The thicker wire must have the smaller slope. The photograph contains no
micrometer readings or measured \((I,V)\) pairs, so it supports this symbolic
answer but no numerical slope or diameter ratio.

\smallskip
\textcolor{casebad}{\textbf{Qwen3VL-8B-Thinking.}}
\textbf{1.} The rollout reverses the axes and defines
\[
\caseerror{S_X=\frac{\Delta I_X}{\Delta V_X}=\frac{1}{R_X}}.
\]
\textbf{2.} It treats the drawn line widths as measurements and sets
\(\caseerror{d_B=2d_A}\).
\textbf{3.} Using those two unsupported premises, it computes
\[
\frac{S_B}{S_A}
=\left(\frac{\caseerror{d_B}}{d_A}\right)^2=4
\]
and reports
\[
\boxed{\caseerror{S_B=4S_A}}.
\]
The result follows only after reversing the plotted axes and measuring a
schematic as though it were drawn to scale.
\end{minipage}

\vspace{0.45em}
\hrule
\vspace{0.45em}

\begin{minipage}[t]{0.31\textwidth}
\vspace{0pt}
\includegraphics[width=\linewidth]{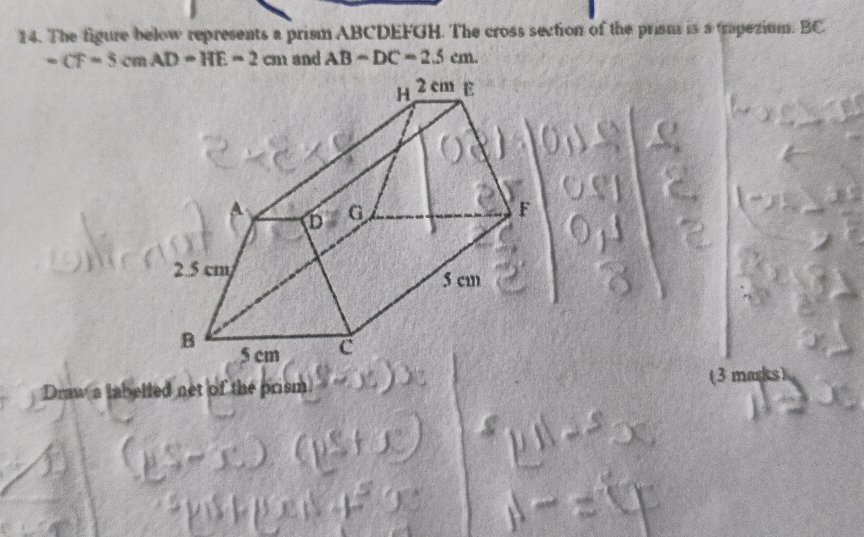}
\end{minipage}\hfill
\begin{minipage}[t]{0.655\textwidth}
\vspace{0pt}
\rolloutformat
\textbf{Net of a trapezoidal prism.}
The two end faces are congruent trapezia. Their side lengths, in cyclic order,
are 2, 2.5, 5, and 2.5 cm. Let \(\ell\) denote the common prism length,
which is not numerically specified in the photographed prompt.

\smallskip
\textcolor{casegood}{\textbf{Our Model.}}
\textbf{1.} Identify the end faces \(ABCD\) and \(HEFG\). They satisfy
\[
AD=HE=2,\quad AB=DC=2.5,\quad BC=GF=5\ \text{cm}.
\]
\textbf{2.} A prism has one lateral rectangle for every side of its
cross-section. The four rectangles therefore have dimensions
\[
2\times\ell,\quad 2.5\times\ell,\quad
5\times\ell,\quad 2.5\times\ell.
\]
\textbf{3.} Place the rectangles in the same cyclic order as the four
trapezium edges. Attach one congruent trapezium to an outer edge of the strip
and the second to the corresponding edge on the opposite side.
\textbf{4.} The labelled net is therefore
\[
\boxed{\text{two congruent trapezia and four lateral rectangles}}.
\]
Several planar arrangements are valid, but every valid net has this face
inventory and preserves the vertex correspondences.

\smallskip
\textcolor{casebad}{\textbf{Qwen3VL-8B-Thinking.}}
\textbf{1.} The rollout treats the 5-cm base label as the prism length and
then assumes \(\caseerror{\text{every edge is }5\text{ cm}}\).
\textbf{2.} It replaces the stated trapezium with
\(\caseerror{\text{a }5\text{ cm}\times5\text{ cm square}}\).
\textbf{3.} It consequently classifies the solid as
\(\caseerror{\text{a cube}}\).
\textbf{4.} Its final net is
\[
\boxed{\caseerror{\text{six }5\text{ cm}\times5\text{ cm squares}}}.
\]
This rollout discards the visible 2-cm and 2.5-cm labels and omits both
trapezoidal end faces.
\end{minipage}
\caption{\textbf{Science and geometry rollouts.} Our Model uses the stated
axes, dimensions, and topology; Qwen3VL-8B-Thinking propagates a specific
diagram-reading error to the final answer.}
\label{fig:case_studies_stem}
\end{figure*}
\FloatBarrier

\section{Scope, limitations, and broader impact}
\label{app:limitations}

VTS is a controlled rendering protocol. The held-out natural-page evaluation
tests generalization across four real-world layout categories, but its unpaired
design does not measure a natural-page semantic-channel gap. It also does not
cover all languages, handwriting, severe occlusion, dynamic interfaces, or
unknown task regions. VISTA-Bench supplies an independent matched
visualized-text protocol, whereas OCRBench-style and document benchmarks test
broader text-rich capability rather than semantic-channel equivalence.

Prompt-region supervision assumes a known task region during training. The
deployed model needs neither a region nor OCR, but the method is not itself a
general task-region detector. The second-backbone result supports transfer to
InternVL3.5-8B under the tested setup; broader architecture and scale studies
remain necessary.

Stronger visual-instruction following creates an additional trust boundary.
Our bounded safety check does not show higher visible or obfuscated attack
success, but it is not a substitute for adversarial evaluation. Systems should
preserve explicit instruction priority, distinguish trusted from untrusted
image content, and support refusal when channels conflict. Any release of
user-contributed screenshots or photographs also requires consent,
de-identification, source and license records, duplicate checks, and a removal
process.

Finally, attention heatmaps and representation retrievals are descriptive.
They can reject simple explanations such as complete representation collapse,
but they do not identify a unique internal mechanism. Our mechanistic wording
is therefore limited to the measured spatial, semantic-target, and behavioral
contrasts.

\end{document}